\documentclass[10pt,twocolumn,letterpaper]{article}

\usepackage[pagenumbers]{cvpr} 

\usepackage[dvipsnames]{xcolor}

\usepackage{color}
\definecolor{green}{rgb}{0, 0.5, 0}
\definecolor{deepred}{rgb}{0.9, 0.15, 0.15}
\definecolor{orange}{rgb}{0.8, 0.6, 0.2}
\definecolor{red}{rgb}{1.0, 0.0, 0.0}
\definecolor{teal}{rgb}{0.0, 0.4, 0.4}
\definecolor{purple}{rgb}{0.65,0,0.65}
\definecolor{saffron}{rgb}{0.75,0.05,0.05}
\definecolor{turquoise}{rgb}{0.0,0.30,0.15}
\definecolor{black}{rgb}{0.0, 0.0, 0.0}
\definecolor{gray}{rgb}{0.5, 0.5, 0.5}

\definecolor{cvprblue}{rgb}{0.21,0.49,0.74}
\usepackage[pagebackref,breaklinks,colorlinks,citecolor=cvprblue]{hyperref}
\usepackage{graphicx}

\def\paperID{3500} 
\def\confName{CVPR}
\def\confYear{2024}

\title{CLiC: Concept Learning in Context}

\author{Mehdi Safaee$^{1}$ \hspace{6mm}
Aryan Mikaeili$^{1}$ \hspace{6mm}
Or Patashnik$^{2}$ \hspace{6mm}
Daniel Cohen-Or$^{2}$\hspace{6mm}
Ali Mahdavi-Amiri$^{1}$\\\\
$^1$Simon Fraser University \hspace{0.25cm}
$^2$Tel Aviv University
\vspace{4mm}
}
\usepackage{cuted}
\usepackage{capt-of}
\usepackage{bm}
\usepackage{svg}

\begin{document}


\twocolumn[{%
\maketitle
\renewcommand\twocolumn[1][]{#1}%
\begin{center}
    \centering
    \vspace*{-1.0cm}
    \includegraphics[width=0.93\linewidth]{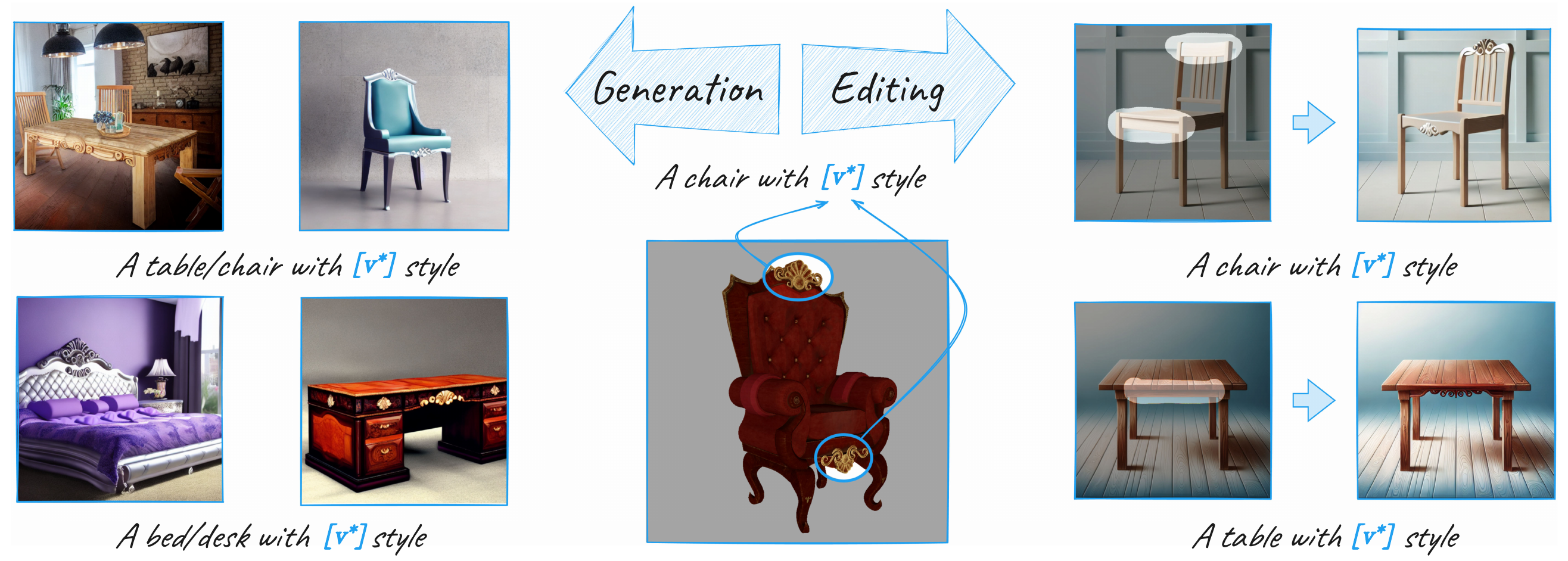} 
    \vspace{-0.2cm}
    \captionsetup{type=figure}\caption{
    Given an object in an image (e.g., the red chair in the middle), we learn an in-context token for a concept embedded in this object (e.g., an ornament of a chair). This token can then be used to generate images with new objects embodying the same concept (left) or to transfer the concept to given target objects, while maintaining their structure (right). Project page: \href{https://mehdi0xc.github.io/clic}{https://mehdi0xc.github.io/clic}
    }
    \label{fig:teaser}
\end{center}
}]

\begin{abstract}

\vspace{-11pt}

This paper addresses the challenge of learning a local visual pattern of an object from one image, and generating images depicting objects with that pattern.
Learning a localized concept and placing it on an object in a target image is a nontrivial task, as the objects may have different orientations and shapes.
Our approach builds upon recent advancements in visual concept learning. It involves acquiring a visual concept (e.g., an ornament) from a source image and subsequently applying it to an object (e.g., a chair) in a target image.
Our key idea is to perform in-context concept learning, acquiring the local visual concept within the broader context of the objects they belong to. 
To localize the concept learning, we employ soft masks that contain both the concept within the mask and the surrounding image area. 
We demonstrate our approach through object generation within an image, showcasing plausible embedding of in-context learned concepts.
We also introduce methods for directing acquired concepts to specific locations within target images, employing cross-attention mechanisms, and establishing correspondences between source and target objects. 
The effectiveness of our method is demonstrated through quantitative and qualitative experiments, along with comparisons against baseline techniques.

\end{abstract}

\section{Introduction}
\label{sec:introduction}

Consider the problem of transferring an ornament from one image of a chair onto another image of a different chair, even if the chairs are in different orientations (see Fig.~\ref{fig:teaser}). It is evident that a straightforward image-space cut-and-paste operation is insufficient here. Moreover, attempting to model the ornament from a single perspective and accurately pasting it onto the other image is a complex task, one that currently stands as a daunting challenge.

An alternative approach involves harnessing the recently developed domain of visual concept learning \cite{gal2022textual,ruiz2022dreambooth,kumari2022customdiffusion,gal2023encoderbased,alaluf2023neural} that allows learning the visual concept from a source image and subsequently applying it to a target image. While it does not provide an exact, one-to-one transfer of the ornament, it does offer a way to transfer the overall concept. Yet, plausibly learning a local concept from a single image and applying it in a specific location of an object in the target image is challenging due to the lack of context (see Fig.~\ref{fig:teaser}).

Avrahami et al. \cite{avrahami2023break} have recently presented a technique called ``Break-A-Scene''  wherein they learn local concepts from a single image and then apply them within a generated image through text-to-image machinery. This technique can be seemingly applied to our local concept learning problem as well. However, as we shall show in the following, our specific task necessitates learning local visual concepts within the shape's context rather than in isolation. Our learned local concepts are intrinsically tied to the objects in which they are embedded. The method we present in this paper addresses the intricate challenge of in-context concept learning, specifically tailored to our requirements.

\begin{figure}
    \centering
    \includegraphics[width=0.99\linewidth]{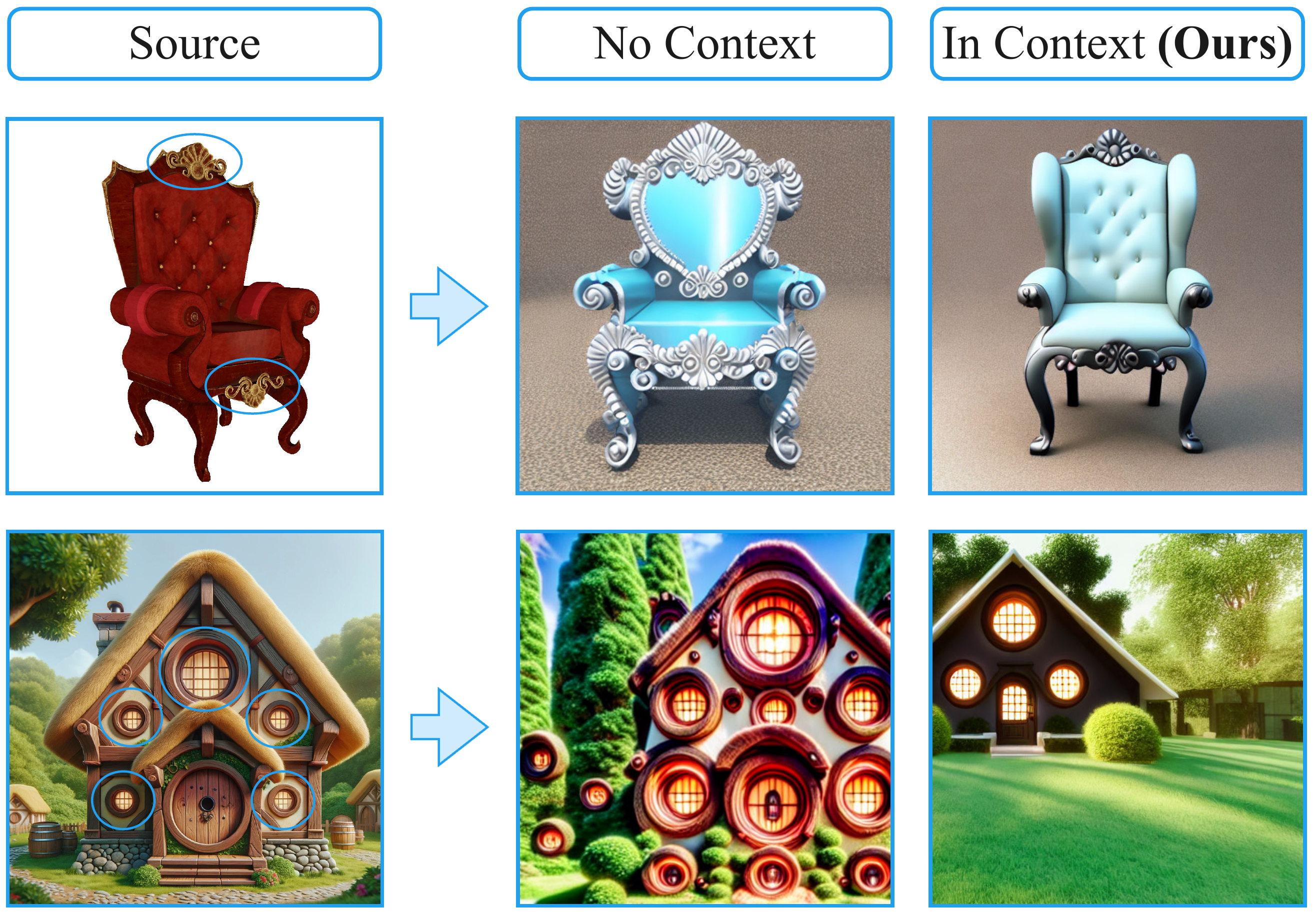}
    \caption{By learning a concept in-context, the comprehension of the concept extends beyond its visual attributes to encompass its relationship with the surrounding context. In this example, when learning the ornaments and windows in-context, they are placed in similar semantic locations in the generated images as in the source image. Conversely, when ornaments are learned without the context, they may be dispersed randomly across the chair and house.}
    \label{fig:motivation}
\end{figure}

To learn a visual concept in-context, we apply a personalization method that learns a token $v^*$, where a mask defines the spatial region of the acquired concept (e.g., ornament/window). Rather than applying the losses only under the given mask, we compute a loss with a non-binary mask, that is, a soft mask that considers both the in-mask concept and the out-mask portion of the image. Fig.~\ref{fig:motivation} shows two examples generated with a simple text prompt, ``A chair/house with $v^*$ style'' where one is with the in-context learned concept (the ornament/windows in the figure), and one without. As can be seen, the in-context visual concept learning successfully embeds the concept only in the expected region of the generated results. 

We show that the acquired concept can be directed to a specific location within a given target image through the optimization of cross-attention layers and the establishment of correspondences between source and target objects. We further present an automated process for identifying common concepts when multiple objects embodying a particular concept are available, which removes the necessity to manually choose the concept in the source image.

We demonstrate the efficacy of the method via numerous results and multiple quantitative and qualitative experiments and comparisons against baseline methods. We also show the necessity of having each component of the method through a series of thorough ablation studies. 


\section{Related Work}

The field of text-conditioned image generation \cite{rombach2021highresolution, imagen2022saharia, ramesh2022dalle, balaji2022eDiff-I} has recently advanced significantly by combining the power of diffusion models \cite{ho2020ddpm, sohl2015diffusion, song2020denoising, song2020improvedsd} and large-scale text-image datasets \cite{schuhmann2022laion5b}.
These advancements have had great contributions to the area of content creation, showcasing the capacity of these models to produce captivating visual content, enabling a multitude of creative visual tasks through image generation and editing \cite{hertz2022prompt, mokady2022null, tumanyan2023pnp, kawar2023imagic, brooks2022instructpix2pix, meng2022sdedit}. One such task is to utilize user-defined concepts~\cite{gal2022textual,ruiz2022dreambooth} to accommodate \emph{personalization} \cite{han2023svdiff, kumari2022customdiffusion, oft2023, alaluf2023neural, voynov2023P+, jia2023taming, wei2023elite}, empowering users to craft expressive content that seamlessly blend subjects and artistic styles, often requiring just a small collection of concept-exemplifying images.

The first attempts to address personalization were Textual Inversion \cite{gal2022textual} and DreamBooth~\cite{ruiz2022dreambooth}. In both works, given multiple images of a single concept, a text token dedicated to that concept is learned. However, while the former freezes the weights of the diffusion model UNet, the latter optimizes the UNet, showing better reconstruction and generalization capabilities at the expense of additional time and memory consumption. 
Custom Diffusion \cite{kumari2022customdiffusion} approaches this problem by optimizing only the cross-attention layers of the diffusion UNet, while OFT \cite{oft2023}, LoRA \cite{hu2021lora}, and SVDiff \cite{han2023svdiff} restrict the parameter updating for more efficient and well-behaved optimization. 
Similarly, PerFusion \cite{tewel2023keylocked} introduces a key-locking mechanism along with rank 1 updating for faster, better, and less memory-consuming personalization. More recently, several works have improved personalization by decreasing runtime and focusing on a single input image \cite{gal2023encoderbased,chen2023suti,shi2023instantbooth,jia2023taming, ruiz2023hyperdreambooth}. Other works learn multiple concepts \cite{avrahami2023break,han2023svdiff,kumari2022customdiffusion} or extend the text embedding space of the diffusion model~\cite{voynov2023P+,alaluf2023neural}. 

In Break-A-Scene \cite{avrahami2023break}, multiple concepts are learned given a single image and user-defined masks. Specifically, the concepts are learned by using a masked diffusion loss and restricting the cross-attention maps of the learned tokens to the input masks. Unlike their work, our method addresses the in-context concept learning, specifically tailored to our requirements. Also, concurrent to our work, RealFill~\cite{tang2023realfill} tackles the problem of personalized image inpainting and outpainting by fine-tuning an inpainting diffusion model~\cite{rombach2021highresolution} on a collection of input images. However, when RealFill is employed for our concept transfer task, the relative size of the concept to the base object is not maintained, and geometric details are compromised. 


\begin{figure*}[t]
    \centering
    \includegraphics[width=0.99\linewidth]{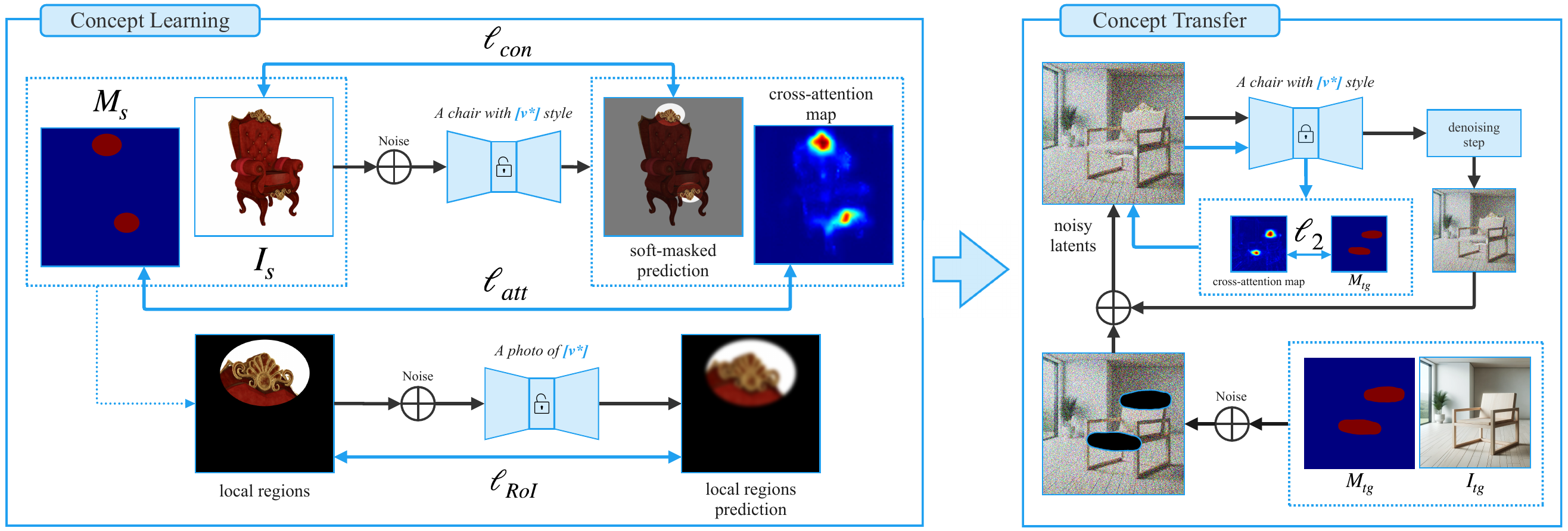}
    \captionsetup{font=small}
    \caption{\textbf{In-Context Concept Learning}: given image $I_s$ and a binary mask $M_s$, we learn $v^*$ for a concept outlined by mask $M_s$ in the context of a base object. Here, the concept is the ornament, and the base object is a chair. Three loss functions are utilized to optimize $v^*$ and fine-tune the diffusion model. $\bm{\ell_{con}}$ uses a soft-masked diffusion loss to learn the pattern in context. $\bm{\ell_{att}}$ ensures that the token focuses only on the pattern region by restricting the attention maps of $v^*$ to $M_s$. By employing a text prompt that is specified for $v^*$, $\bm{\ell_{ROI}}$ enhances the reconstruction of the concept by focusing on a local region through masking $I_s$. \textbf{Concept Transfer}: given image $I_{tg}$, mask $M_{tg}$ defining the area of edit, and a user-defined text-prompt containing $v^*$ optimized in the concept learning step, we add noise to the latent of $I_{tg}$ and denoise it with the fine-tuned diffusion model obtained from the concept learning step. At each denoising step, we blend the output of the diffusion model with the masked input to preserve the out-of-mask regions. We also have a cross-attention guidance to enhance the presence of the pattern in the final output.}
    \label{fig:pipeline}
\end{figure*}

Recently, many works have utilized the intermediate features of diffusion models for image editing \cite{hertz2022prompt, mokady2022null, tumanyan2023pnp, tokenflow2023, cao_2023_masactrl}, controlled image generation \cite{chefer2023attendandexcite, patashnik2023localizing, hong2022improving}, and image understanding \cite{hedlin2023unsupervised, khani2023slime, luo2023dhf, zhang2023tale, tang2023dift}.
Prompt2Prompt \cite{hertz2022prompt} shows that by manipulating the cross-attention layers of the diffusion model, it is possible to edit a certain semantic region of an image. 
Attend-and-Excite \cite{chefer2023attendandexcite} ensures that the diffusion model attends to every subject in the text prompt by manipulating the cross-attention maps of the subjects in the generation process. More recent methods~\cite{hedlin2023unsupervised,khani2023slime} demonstrate that by optimizing text tokens and cross-attention maps, it is possible to establish semantic correspondence or segmentation.  
In our work, we also use the cross-attention maps of the diffusion model to localize the learned tokens to the acquired visual concepts in the source image and automatically place them correctly in the target image.

\section{Method}
\label{sec:method}

Given a source image, denoted as $I_{s}$, a user-specified prompt $P_{s}$, and a learned or user-provided mask $M_{s}$ that marks the Region of Interest (RoI) within the source image, our objective is to learn the concept (e.g., an ornament) in the RoI. To achieve this objective, we build on previous works that employed text-to-image diffusion models for the task of personalization.
Such works either optimize a text token $v^*$, fine-tune the pretrained text-to-image model, or their combination.
In our work, we opt to use Custom Diffusion~\cite{kumari2022customdiffusion}, optimizing a text token $v^*$ and simultaneously fine-tuning the cross-attention layers of the text-to-image model.
To learn the concept in-context (Section~\ref{sec:inversion}), we employ
multiple loss functions, encouraging the diffusion model to reconstruct the learned concept in analogous contexts but under varying conditions and poses.
After learning the concept, we can either generate images that contain it, or edit a target image $I_{tg}$ to portray it in a given RoI.
The RoI in the target image is determined either via our Diffusion-Based RoI Matching Algorithm proposed in Section~\ref{sec:RoI Matching} or directly provided by the user. 

\subsection{In-Context Concept Learning}
\label{sec:inversion}
To acquire in-context concepts from the source image, we optimize token $v^*$ and simultaneously fine-tune the cross-attention layers of a pretrained T2I diffusion model as done in Custom Diffusion~\cite{kumari2022customdiffusion}.
We employ three loss functions to ensure effective concept learning and precise in-context generation. $\ell_{att}$ helps the model to focus on the RoI. $\ell_{context}$ facilitates in-context concept learning. Although the concept is learned in-context, ensuring that the token possesses knowledge of the concept's inclusion within a larger object, we employ $\ell_{RoI}$ to safeguard against overfitting the concept to a particular object in the source image. This approach enhances the concept's ability to generalize and transfer to unseen objects, even those from different categories. Additionally, $\ell_{RoI}$ aids in acquiring a more nuanced understanding of the concept's geometric attributes.

Given the source image $I_{s}$, its corresponding binary mask $M_{s}$, and a text prompt $P_{s}$, we first encode the image and prompt to obtain latent image $x_0$ and text embedding $c$. Thereafter, by randomly sampling a timestep $t$ from the interval $[1, T]$ and a noise $\epsilon$, we construct a noisy latent $x_t$. We then employ the diffusion model to get $\epsilon_{\theta}(x_t, c, t)$ while also extracting cross-attention maps for the token $v^*$ from the decoder layers of the UNet structure. The cross-attention loss is then computed as:
\begin{equation}
   \ell_{att} = \mathbb{E}_{\left(x_t, t\right)}\left[\left\|C A_{\theta}\left(v^*, x_t\right) - Resize(M_{s})\right\|_2^2\right],
   \label{eq:attn_loss}
\end{equation}
where $C A_{\theta}\left(v^*, x_t\right)$ denotes cross-attention maps between token $v^*$ and $x_t$ averaged over the cross-attention layers of the upsampling blocks and $Resize(M_{s})$ is the resized version of $M_{s}$ that matches the shape of cross-attention maps.

For the context loss, we aim to make the model focus on in-context reconstruction of the concept in the RoI while simultaneously forcing the model to focus on the proper scale and placement of this concept. To achieve this, we employ a soft-weighted version of $M_{s}$:
\begin{equation}
M_{soft}=\alpha+(1-\alpha) M_s,
\label{eq:context_loss}
\end{equation}
where $\alpha=0.5$. The context loss is then computed as:
\begin{equation}
   \ell_{con}=\mathbb{E}_{\left(x_t, c, t\right)}\left[\left\| M_{soft} \odot (\epsilon_\theta\left(x_t, c, t\right)-\epsilon)\right\|_2^2\right].
\end{equation}
For RoI loss, we use a more concept-oriented prompt ``A photo of $v^*$'' encoded into $c^*$:
\begin{equation}
\ell_{R o I}=\mathbb{E}_{\left(x_t, t\right)}\left[\left\|\epsilon_\theta\left(M_{s} \odot x_t, c^*, t\right)-\epsilon\right\|_2^2\right]
\label{eq:roi_loss}
\end{equation}
 and we finally add all the losses and perform optimization:
\begin{equation}
\ell_{tot}=\ell_{con} + \lambda_{att} \ell_{att} + \lambda_{RoI} \ell_{RoI},
\label{eq:total_loss}
\end{equation}
where $\lambda_{att}$ and $\lambda_{RoI}$ are empirically set to $0.5$.

\subsection{Concept Transfer}
\label{sec:transfer}


To transfer the learned concept to new objects while preserving the region outside the Region of Interest (RoI), we utilize masked blended diffusion editing~\cite{avrahami2020blendeddiffusion}. This involves adding a specific amount of Gaussian noise to the target image to reach the timestep $t_{start}$ of the denoising process. We then begin denoising the image, simultaneously blending the out-of-mask region of the target image at each denoising step. Additionally, adapted from Attend-and-Excite~\cite{chefer2023attendandexcite}, we introduce cross-attention guidance to improve control over the strength of the edit. In this process, we gradually optimize the latents so that the cross-attention map of the $v^*$ token increases in the RoI and decreases elsewhere.

\vspace{-12pt}
\paragraph{Blended Diffusion Editing.} Given a target image $I_{tg}$ and its corresponding mask $M_{tg}$, we aim to modify segments within $ M_{tg} \odot I_{tg}$. First, the target image is encoded, and $x_{tg}$ is obtained, then an initial timestep $t_{start}$ is chosen ($5 \leq t_{start} \leq 15$). Next, we add $T-{t_{start}}$ levels of noise to $x_{tg}$ to obtain  $x'_{t_{start}}$, then, at each timestep $0 < t\leq t_{start}$, blended output $x'_t$ is computed as:

\begin{equation}
x_t^{\prime}= M_{tg} \odot x_t + \left(1-M_{tg}\right) \odot x'_{start}.
\label{eq:blending}
\end{equation}

\paragraph{Cross-Attention Guidance.} After obtaining $x_t^{\prime}$, we extract the attention maps of the $v^*$ token $CA_\theta (v^*, x'_t)$. Then, we update $x'_t$ according to Equation~\ref{eq:optimization} to enhance the strength of the attention maps of $v^*$ within $M_{tg}$:
\begin{equation}
x_t^{\prime \prime}=x_t^{\prime}-\eta \nabla\mathbb{E}_{l}[\left\|C A_{\theta}\left(v^*, x_t^{\prime}\right)-M_{tg}\right\|_2^2].
\label{eq:optimization}
\end{equation}
Here $\eta$ is the step size of the guidance. changing this parameter controls the strength of the edit in the RoI. Finally, we denoise $x_t^{\prime\prime}$ through the UNet.

\begin{figure}
    \centering
    \includegraphics[width=0.99\linewidth]{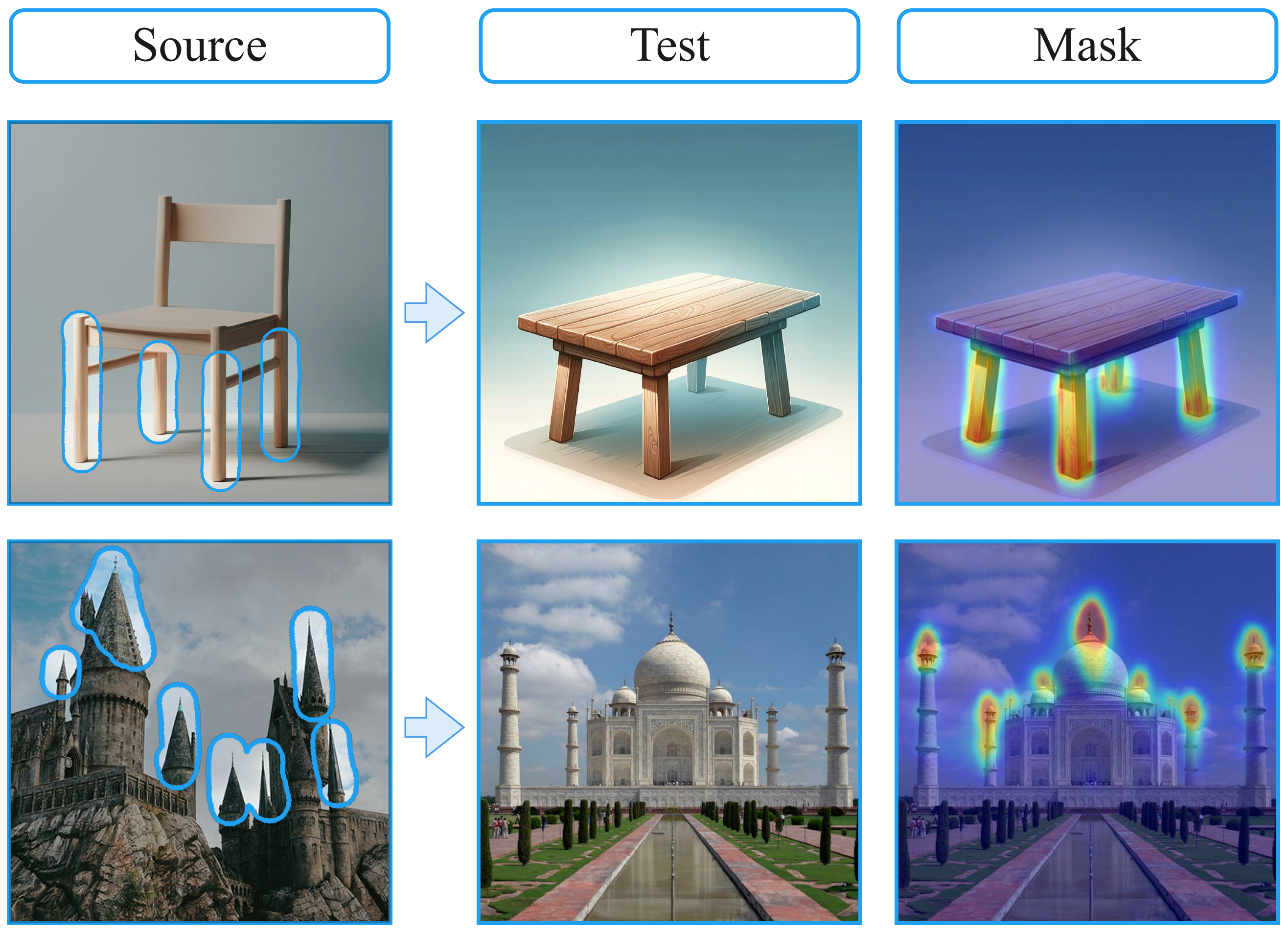}
    \caption{Illustration of the automated masking process on target images employing the proposed RoI-Matching technique, leveraging a predefined source mask.}
    \label{fig:correspondece}
\end{figure}

\subsection{RoI Matching}
\label{sec:RoI Matching}
\textbf{Automatic Target Mask Extraction}. Mask extraction on the target image according to the source input mask can be automated. The idea is to learn a new token $w^*$ to the text encoder, initialized with the already optimized $v^*$ and optimizing it by minimizing our attention loss, $\ell_{att}$, using the prompt ``a $w^*$ region of an OBJECT'', with OBJECT being the base object in the source image. After 500 steps of optimization, we apply the new token on the target or other source images and execute the denoising process, extracting the attention maps of the token $w^*$ as the target masks. By doing this, the model 
acts as a segmentation method that segments the corresponding part of the target or source images.  In Fig.~\ref{fig:correspondece}, we demonstrate that this automatic masking technique works well both for in-domain and cross-domain scenarios. Notably, this process is quite fast since we only fine-tune the newly added token. 
\begin{figure}
    \centering
    \includegraphics[width=0.99\linewidth]{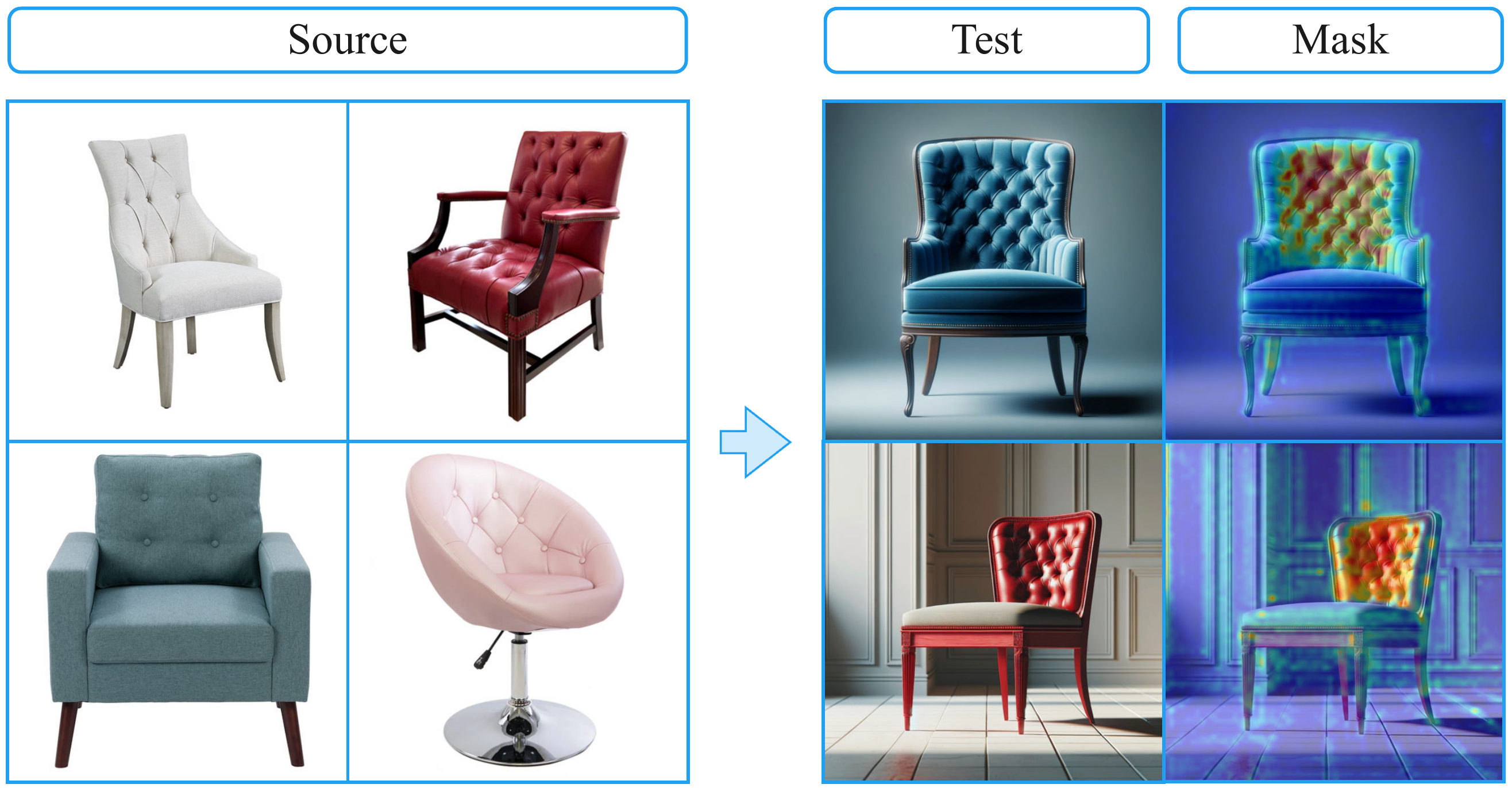}
    \caption{ Given several images with a common pattern, our method is able to learn the common pattern and locate it even on a different image with the same pattern.
    }
    \label{fig:automation}
\end{figure}
\vspace{-12pt}
\paragraph{Automatic Source Mask Extraction.} When multiple source images sharing the desired concept exist, it can be automatically identified. To do so, we add a token $w^*$ to the text encoder and optimize its embedding and the cross-attention modules of the diffusion UNet by minimizing the diffusion loss, given the prompt ``An OBJECT with $w^*$ style", OBJECT being the class of base object containing the pattern. After 500 steps, we extract the attention maps of $w^*$ and use them as the source mask, and run our concept learning pipeline. This method is effective when multiple images of an object containing the concept exist (Fig.~\ref{fig:automation}) but for unique concepts, it could be simpler to just provide the source mask.
\section{Results and Comparisons}

\begin{figure}
    \centering
    \includegraphics[width=0.95\linewidth]{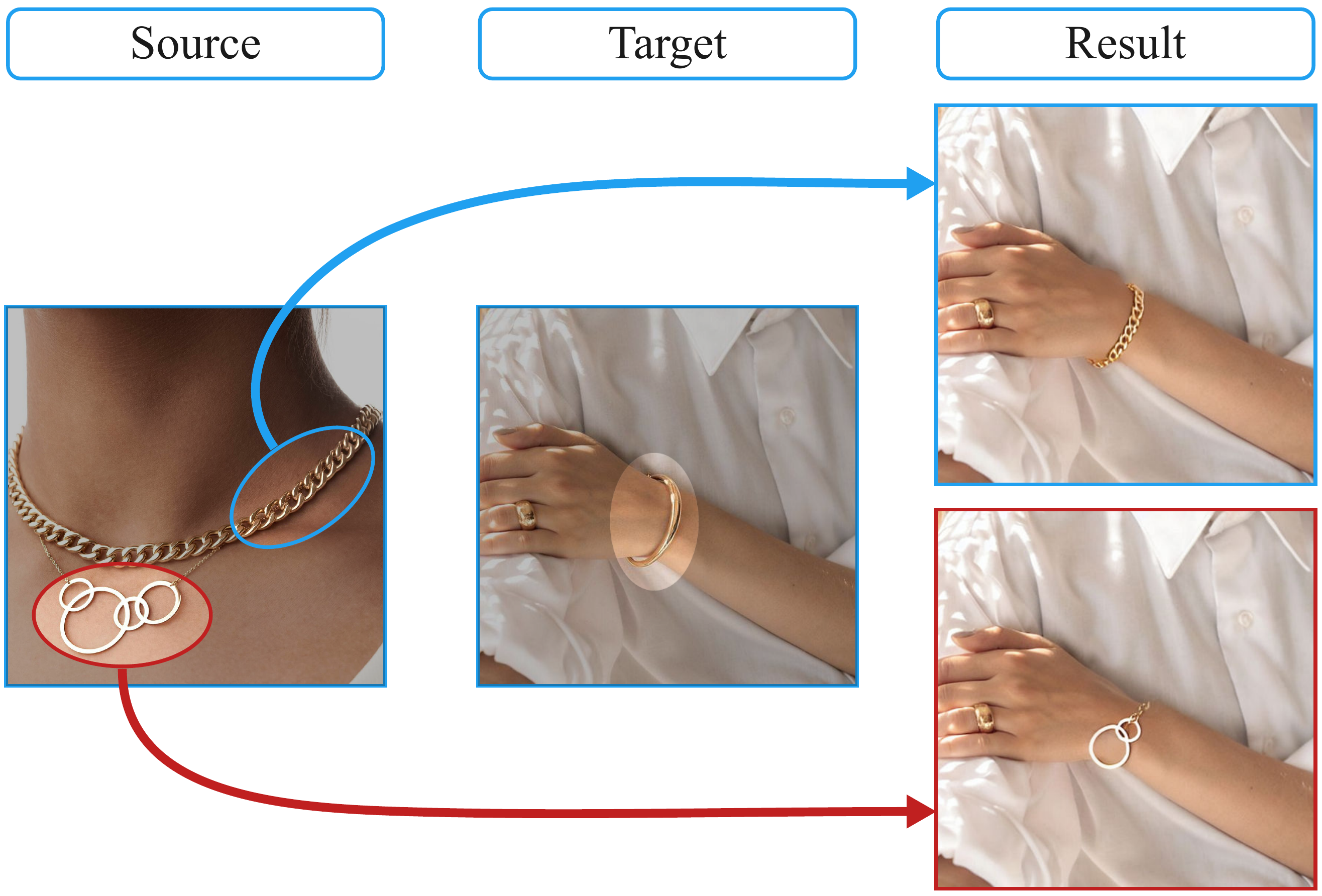}
    \caption{An example of selective pattern extraction: We show that our model can learn distinct patterns from a single image, ensuring that each token captures only its corresponding pattern. Left, we choose two different patterns from a single necklace and transfer them to a bracelet (Right). }
    \label{fig:selective}
\end{figure}

Here, we first qualitatively demonstrate the effectiveness of our method in learning concepts in-context. We show that our learned concepts can be used for generation and transferred across images.
We compare our method with multiple customization methods such as Custom Diffusion \cite{kumari2022customdiffusion}, Break-A-Scene \cite{avrahami2023break}, and RealFill \cite{tang2023realfill}, showing the superiority of our method (Section~\ref{sec:comparison}). We also provide a user study, confirming the effectiveness of our method compared to these baselines. Finally, we ablate the components of our pipeline to justify our design (Section~\ref{sec:ablation}).

In all of our experiments, we use StableDiffusion v1.4 from the diffusers library \cite{diffusers}. We run our in-context concept learning for 500 steps, taking approximately 3 minutes on a single Nvidia RTX3090 GPU. We use the Adam~\cite{kingma2017adam} optimizer with learning rate $1e^{-5}$.

\subsection{Qualitative Results}
\label{sec:qual}
\begin{figure*}[t]
    \centering
    \includegraphics[width=1\linewidth]{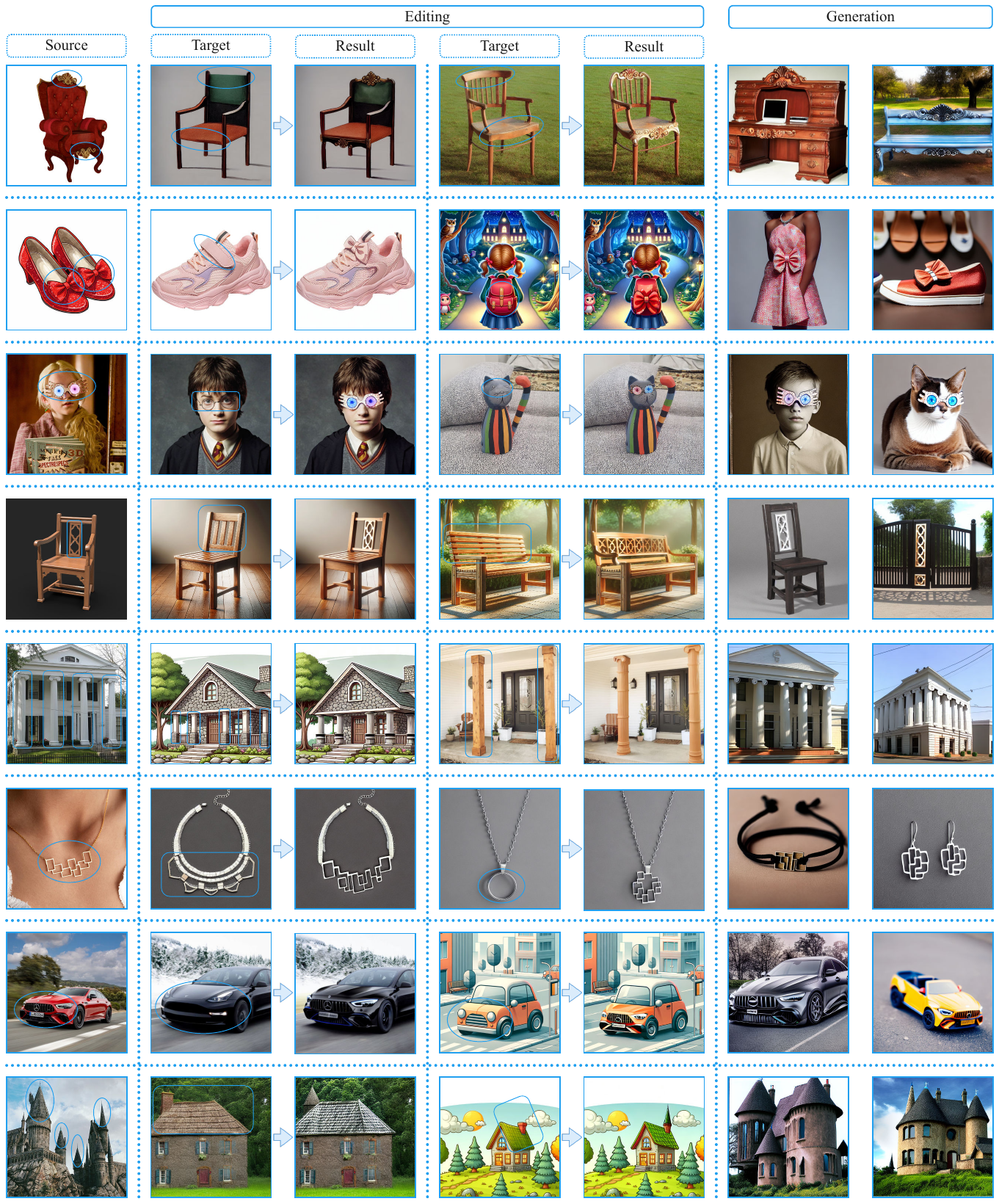}
    \captionsetup{font=small}
    \caption{Results of our concept transfer (Middle) and generation (Right). 
    Concepts delineated by blue curves in the source image are learned and transferred to target images at the locations indicated by blue curves (Middle). The same concepts are used to generate various objects in each row (Right). Our method is successful in learning the concept and placing it coherently within the target or generated image.}
    \label{fig:results_editing}
\end{figure*}

\paragraph{Editing.} 
Our method can successfully learn various concepts and transfer them to objects of the same or different class in an image.
Fig.~\ref{fig:selective} shows that our method can learn individual concepts from a single image 
without color and shape information from other concepts leaking into the token. Fig.~\ref{fig:results_editing} illustrates examples of various classes.
Note that the learned concepts are blended nicely in the target image, attaining the target's texture and color style even when the target domain is very different from the source domain, such as the cartoonish car and house examples.
This demonstrates that our approach does not suffer from overfitting to the concept or the content of the source image, and it reaffirms that a simple copy-and-paste algorithm is not suitable for achieving our objective.

\vspace{-12pt}
\paragraph{Generation.} To generate an object containing the learned concept, we employ a two-stage generation strategy. Starting from a Gaussian noise, for the first $t_s = 5$ steps of denoising, we use the un-modified UNet with the text prompt ``a photo of an OBJECT" where OBJECT is the object we want to generate. After the $t_s$ steps, we substitute the UNet with our fine-tuned model and use the text prompt ``a photo of an OBJECT, with $v^*$ style".

This way, we leverage the capabilities of the pre-trained diffusion model in generating general realistic images while integrating specific patterns and concept details into the output through our fine-tuned model, which possesses an enhanced understanding of our desired concept. 
We present our generation results in Fig.~\ref{fig:results_editing}. Observe that concepts learned from an object (e.g., chair), can be effectively used to generate other objects embodying the same concepts.

 \begin{figure*}[t]
    \centering
    \includegraphics[width=0.98\linewidth]{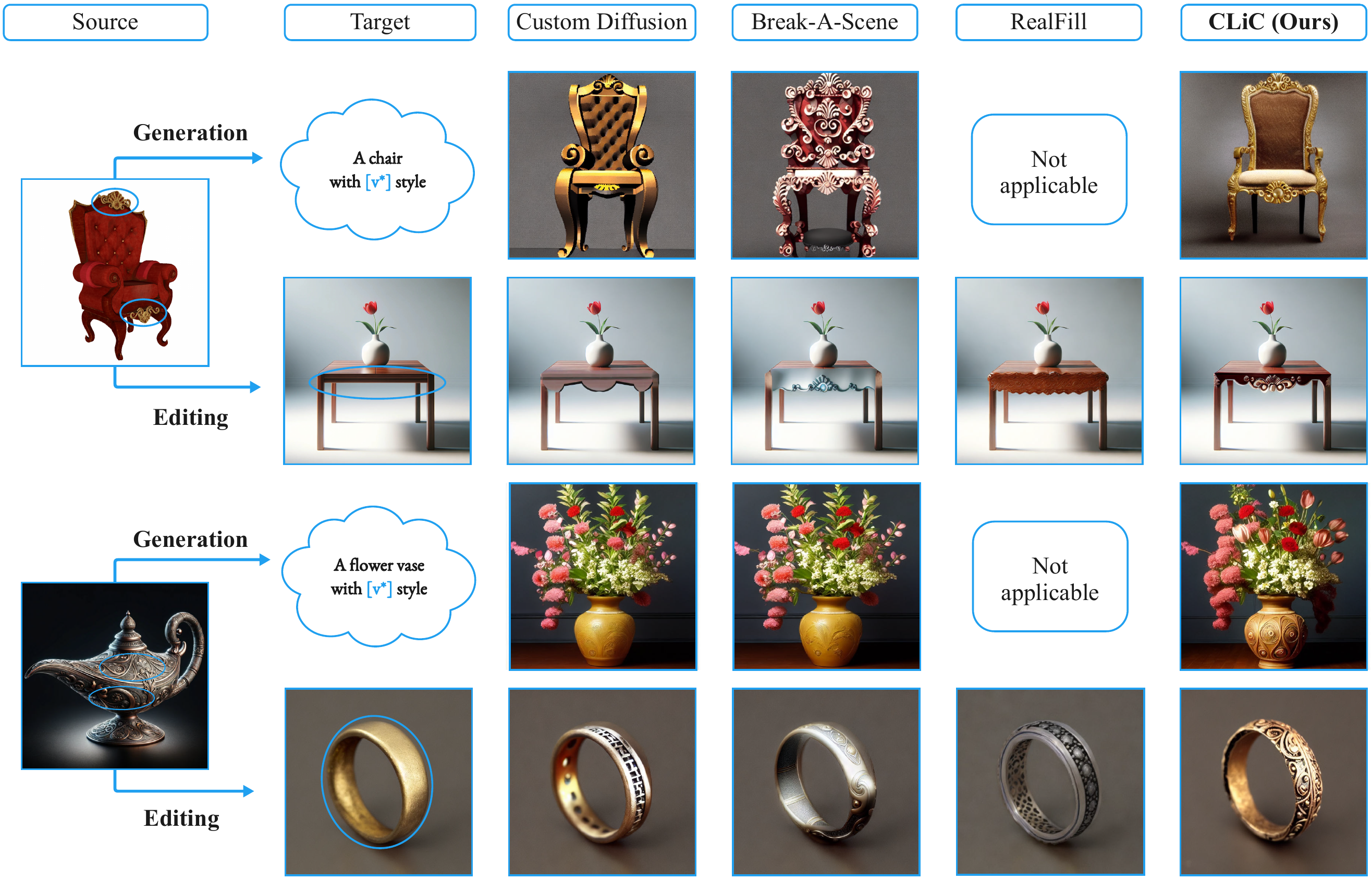}
    \captionsetup{font=small}
    \caption{\textbf{Comparisons}. Given a source image and concepts of interest (Left), the task is either to generate an object (written in bold (Top)) with that concept or transfer the concept to a target object in another image (Bottom). In comparison with alternative methods, our method clearly remains more faithful to the concept in terms of structure and geometric features in both generation and editing.}
    \label{fig:comparison}
    \vspace{-16pt}
\end{figure*}

\subsection{Comparisons}
\label{sec:comparison}

We compare our method against several personalization baseline methods, including Custom Diffusion~\cite{kumari2022customdiffusion}, Break-A-Scene~\cite{avrahami2023break}, and RealFill~\cite{tang2023realfill}.
In Custom Diffusion~\cite{kumari2022customdiffusion}, cross-attention blocks, along with token $v^*$, are optimized for customization by minimizing the unmasked diffusion loss. 
We run Custom Diffusion in a consistent manner with our setting, with inputs consisting of the source image and the text prompt ``an OBJECT with $v^*$ style," where OBJECT denotes the object category (e.g., chair) embodying the concept.
\mbox{Break-A-Scene}~\cite{avrahami2023break} learns several concepts from a single image using masks indicating different subjects. Similar to our setting, we optimize the cross-attention blocks and token $v^*$ representing a local mask located on the concept of interest.
We also conduct a comparison against a concurrent work, RealFill~\cite{tang2023realfill}, designed for personalized inpainting/outpainting. RealFill takes multiple images of a scene as input, randomly applies masks, and refines the Stable Diffusion inpainting model through a process similar to DreamBooth~\cite{ruiz2022dreambooth}. Our transfer task can be viewed as inpainting. To adapt RealFill to our task, we learn the concept delineated by the mask on the source image and optimize the cross-attention blocks of the Inpainting Stable Diffusion. For transfer, we use the fine-tuned UNet and inpaint the masked regions of the target image using the token acquired from the concept present in the source image.

\paragraph{Qualitative Comparison.} In Fig.~\ref{fig:comparison}, we present qualitative comparisons with the baselines. Custom Diffusion struggles to capture the concept present in the source images, failing to transfer the concept to the target images effectively. Break-A-Scene exhibits a relatively good understanding of the concept. However, due to the absence of in-context constraints in the concept-learning process, the model learns the pattern as an independent object, failing to transfer the concept as a pattern. This results in unwanted color and geometry artifacts. Similarly, in RealFill, the model learns the token, yet encounters two challenges. First, using the Stable Diffusion inpainting pipeline results in the loss of information masked by the target mask, preventing the model from preserving the geometry and color of the object in the target image (ring in Fig.~\ref{fig:comparison}). Second, the absence of in-context learning causes the model to fill the entire mask with the pattern without placing it coherently within the target object (table in Fig.~\ref{fig:comparison}).

\vspace{-7pt}
\paragraph{User Study.}
We also conducted a user study using 30 pairs of source and target images. Results of our method and three other baselines (depicted in Fig.~\ref{fig:comparison}), were presented to 42 participants. The 30 images were divided into two sets, each with a consistent number of object classes (buildings, furniture, jewelry, and kitchenware). Participants ranked the methods based on ``edit quality" (accuracy in reflecting the source image concept) and ``target preservation" (maintaining the general appearance and color of the target image). Scores were computed by assigning ranks (4 for the top, 1 for the lowest) and averaging over all samples.
Our method consistently outperformed the three baselines, as detailed in Table~\ref{tab:user_study}. Compared to the second place, RealFill, our method showed a significantly higher score.

\begin{table}
	  \centering
    \captionsetup{font=small}
    \caption{\textbf{User study}. Our method has received a significantly higher score than the alternatives.}
    \small 
	\begin{tabular}{lc}
	\toprule
        Method & Average Ranking ($\uparrow$)\\
		\midrule
        CustomDiffusion~\cite{kumari2022customdiffusion}  & 1.96\\
        Break-A-Scene~\cite{avrahami2023break} & 2.27\\
        RealFill~\cite{tang2023realfill} & 2.33\\
        Ours  & \textbf{3.43}\\
		\bottomrule
	\end{tabular}

    \label{tab:user_study}
\end{table}

\subsection{Ablation Studies}
\label{sec:ablation}

\begin{figure}
    \centering
    \includegraphics[width=0.99\linewidth]{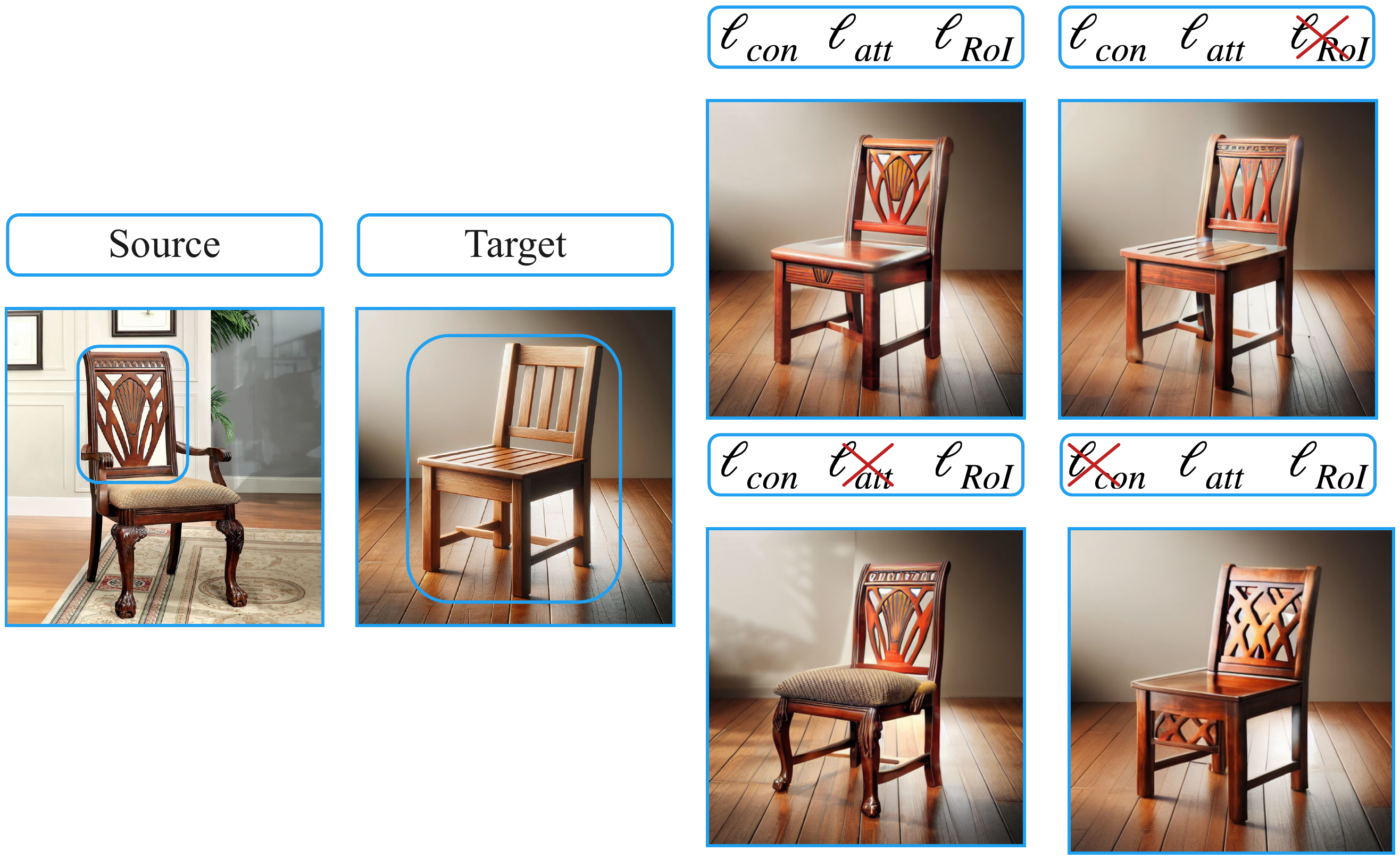}
    \caption{\textbf{Ablation on $\ell_{att}$, $\ell_{RoI}$, $\ell_{con}$}. Omitting $\ell_{RoI}$ causes inaccurate learning of the concept (back seat). Excluding the $\ell_{att}$ produces unintended or off-target edits (seats and legs). Removing $\ell_{con}$ leads to the loss of geometric features and structures associated with the concept (back seat) and also results in the transfer of concepts to undesired regions (the transition between two legs).}
    \label{fig:ablations}
\end{figure}

\paragraph{Loss Ablations.}
Fig.~\ref{fig:ablations} illustrates how each loss in our approach affects the overall performance when the target region encompasses the entire object. Eliminating $\ell_{RoI}$ (top right) leads to the loss of geometric and structural patterns from the source concept, particularly noticeable on the backseat. Removal of $\ell_{att}$ (bottom left) causes undesired alterations on the target, affecting areas such as the legs and seat. The absence of $\ell_{con}$ (bottom right) leads to loss of geometric details and structures associated with the concept (back seat) and results in the unintended transfer of concepts to undesired regions (transition between two legs); same artifacts illustrated in Fig.~\ref{fig:motivation} for the generation process.

\begin{figure}
    \centering
    \includegraphics[width=0.99\linewidth]{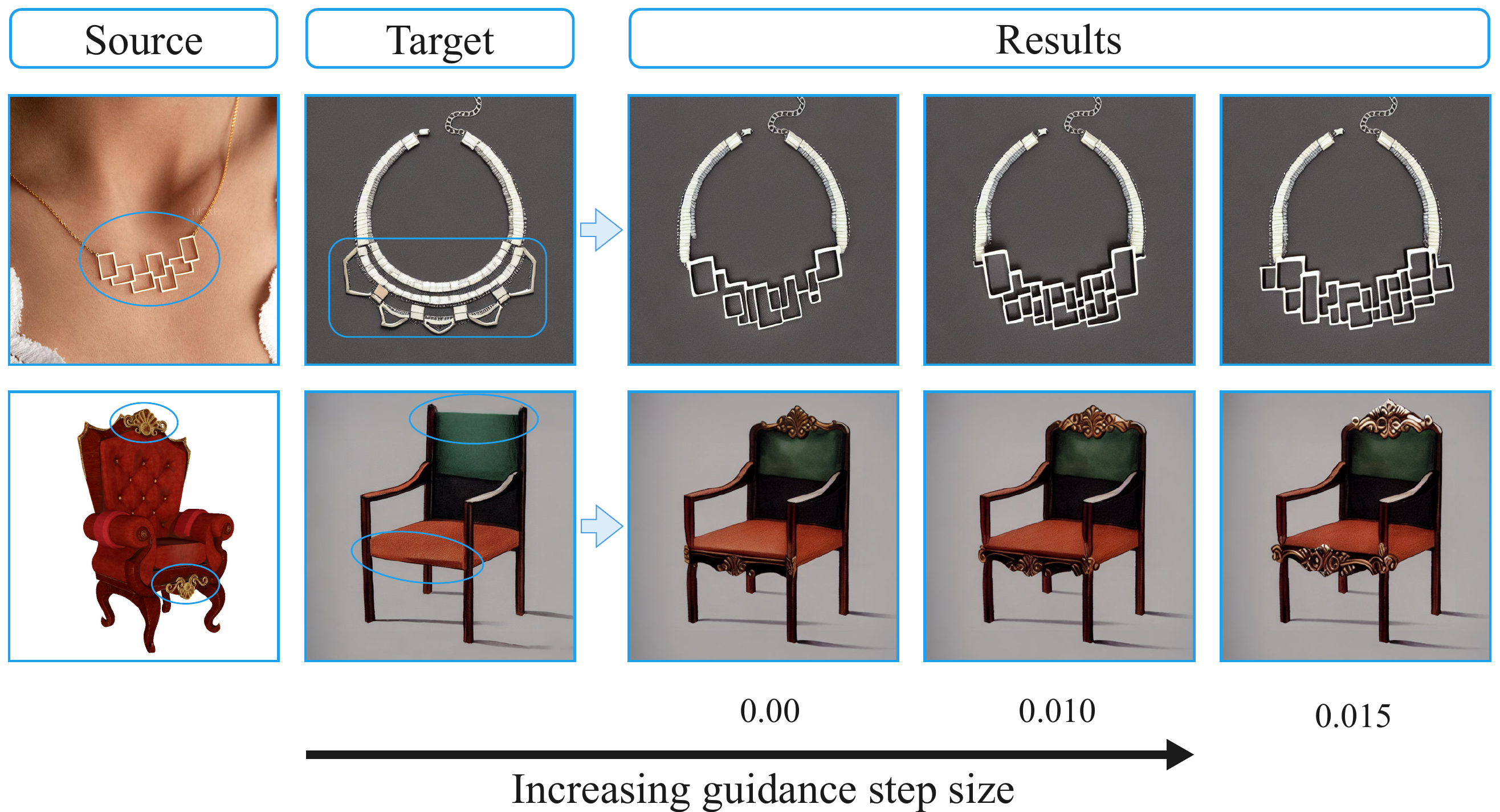}
    \caption{\textbf{Cross-Attention Guidance}. By increasing the guidance step size $\eta$ the presence of the concept is strengthened.}
    \label{fig:guidance}
\end{figure}

\vspace{-12pt}
\paragraph{Cross-Attention Guidance.} As described in Section~\ref{sec:transfer}, we use cross-attention guidance to enhance the presence of the concept to the RoI in the target image while also restricting it to the RoI. In Fig.~\ref{fig:guidance}, one can observe that by changing the guidance step size $\eta$ the presence of the concept in the target image can be adjusted.

\begin{figure}
    \centering
    \includegraphics[width=0.99\linewidth]{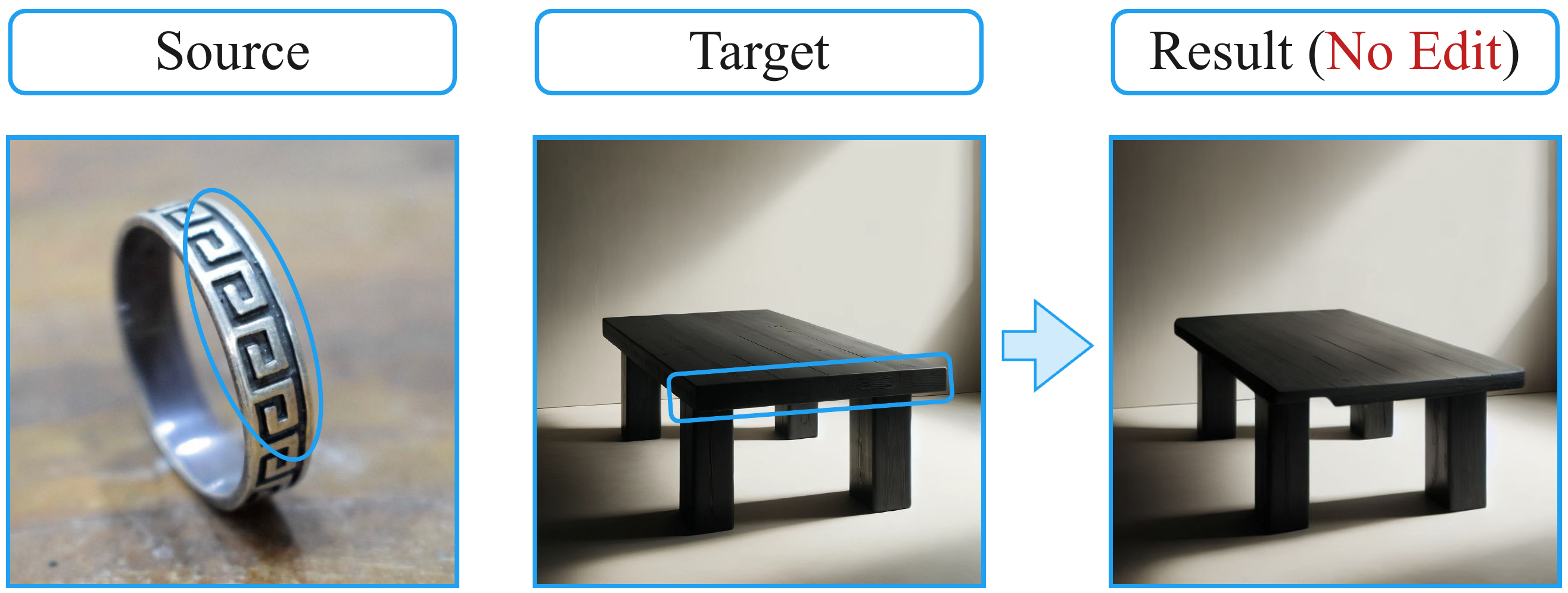}
    \caption{\textbf{Failure Case}. When the domain of the source and target images are too different, concept transfer may fail. }
    \label{fig:failure}
\end{figure}

\section{Discussion and Conclusions}
\label{sec:conclusion}

We have addressed the challenge of learning and transferring visual concepts between images, focusing on acquiring and applying local visual concepts in-context. Traditional cut-and-paste methods have proven insufficient in these contexts, motivating the exploration of visual concept learning.
Our personalization method, which considers both in-mask and out-mask regions of an image, has proven successful in embedding concepts accurately.
Precise concept placement has been achieved through the optimization of cross-attention layers and object correspondences, complemented by an automated concept selection process that streamlines the overall workflow.

\vspace{6pt}

We have demonstrated the efficacy and versatility of our method, and its capability to learn local concepts for editing and generation. However, we acknowledge certain limitations. Our method may exhibit sub-optimal performance when there is a significant difference in the domain of the target image or the objects for generation compared to the source image (see Fig.~\ref{fig:failure}). Additionally, our optimization process, while effective, is time-consuming and not applicable to real-time applications.
We have validated our approach through a diverse set of experiments, quantitative assessment through a user study, a series of qualitative assessments and ablation studies, and comprehensive comparisons with baseline methods.
Exploring the potential of our method for 3D concept transfer and geometry editing presents an intriguing avenue for future research.

\clearpage
{
    \small
    \bibliographystyle{ieeenat_fullname}
    \bibliography{main}
}

\appendix

\clearpage
\setcounter{page}{1}
\maketitlesupplementary

Here, we provide more results, comparisons, and additional implementation details to better prove the efficacy of our technique.
\section{Additional Results}
\label{sec:results}
In this section, we present supplementary results of our method. We begin by offering further examples of our concept transfer and generation method, detailed in Section \ref{sec:qual-sup}. Subsequently, we include additional comparison results against baseline methods in Section \ref{sec:comp-sup}.

\subsection{Qualitative results}
\label{sec:qual-sup}
\begin{figure*}[t]
    \centering
    \includegraphics[width=0.95\linewidth]{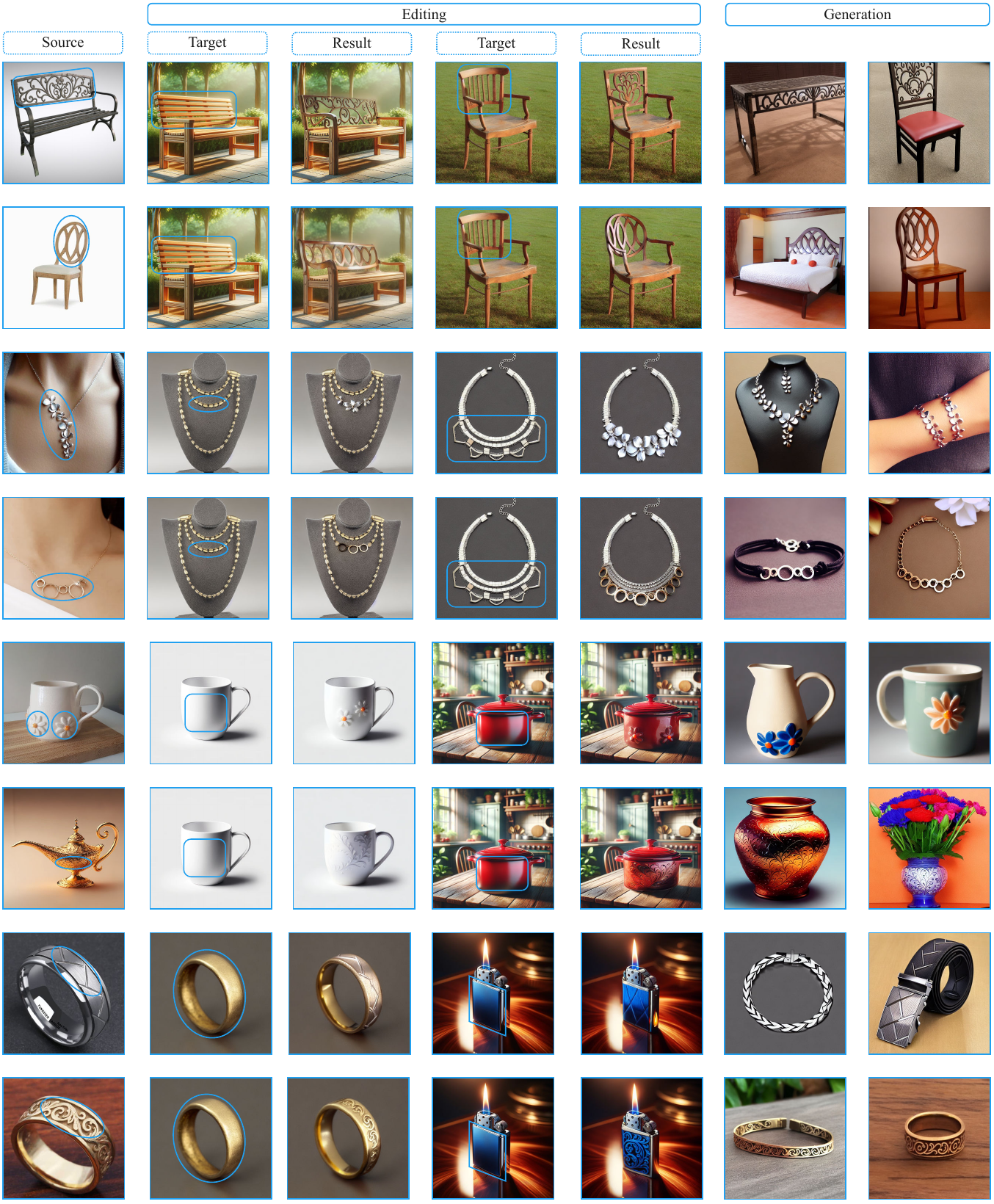}
    \captionsetup{font=small}
    \caption{\textbf{
    Additional editing and generation results}. We have transferred the concept from the source to two targets in each row. We also used the same concept for generation (the last two images in each row).
    }
    \label{fig:results-sup}
    \vspace{-16pt}
\end{figure*}

Fig.~\ref{fig:results-sup} showcases additional results of our concept transfer and generation applications. 
The settings employed for concept transfer and generation are consistent with those outlined in Sections 3.2 and 4.1. Evidently, our method successfully learns concepts from a variety of objects and utilizes these concepts for image editing and generation.

\subsection{Comparison}
\label{sec:comp-sup}
\begin{figure*}[t]
    \centering
    \includegraphics[width=0.95\linewidth]{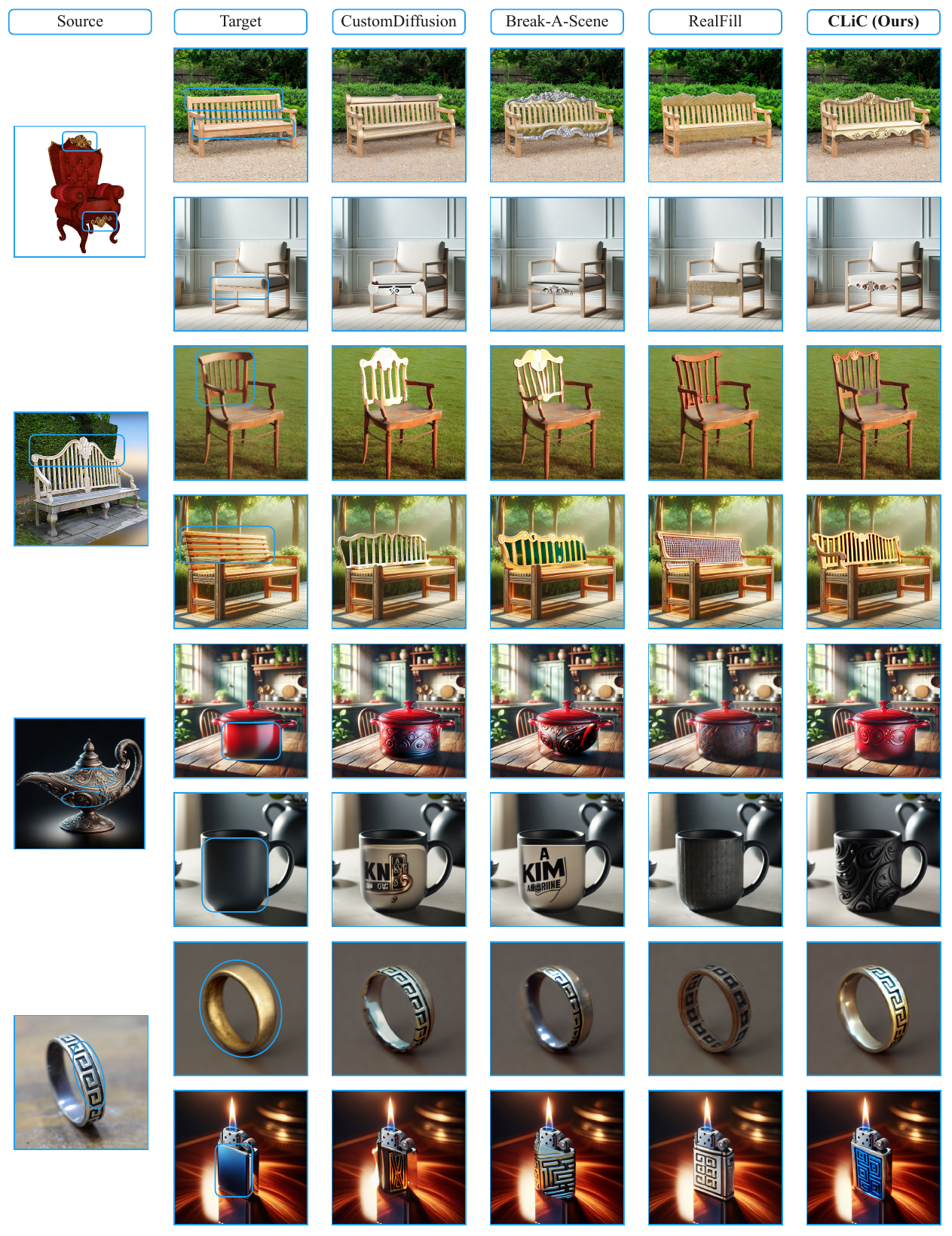}
    \captionsetup{font=small}
    \caption{\textbf{Additional comparisons}. We further compare our concept transfer method with CustomDiffusion~\cite{kumari2022customdiffusion}, Break-A-Scene~\cite{avrahami2023break}, and RealFill~\cite{tang2023realfill}.}
    \label{fig:comparison-sup}
    \vspace{-16pt}
\end{figure*}

In Figure \ref{fig:comparison-sup}, we present additional comparison results alongside the four baselines previously introduced in Section 4.2. It is clearly demonstrated that our in-context concept learning approach exhibits superior proficiency in learning and transferring concepts.

\section{Additional Training Details}

\subsection{Data Augmentation Strategies}
To enhance the robustness of our approach, we incorporated several data augmentation techniques during the training process. These include implementing random grayscaling to reduce dependence on color features, and preventing overfitting to specific colors. We also applied random horizontal flipping to introduce pose diversity, as well as zooming in and out to vary the scale. To address different color intensities and contrasts, we also employed color jittering.

\subsection{Standardized Prompt Templates}
For consistency and to prevent the impact of prompt manipulation, we defined a fixed prompt template and used that for all our experiments. Throughout the Concept Learning phase, we utilized a standardized prompt template: "A {OBJECT} with [v*] style". This uniformity enables effective concept learning and encoding within the [v*] token.

During zoom-in/out data augmentation, the prompt format was dynamically adjusted to reflect these changes. For instance, a zoom-out augmentation led to a prompt alteration to "A {OBJECT} with [v*] style, zoomed-out".

To maintain equitable comparisons, these augmentations and prompt adjustments were consistently applied across all baseline methods.

\subsection{Scheduler Selection}
We opted for the DDIM~\cite{song2020denoising} scheduler for both concept learning and transfer phases, due to its efficiency, speed, and simplicity.
A maximum of 50 timesteps ($T=50$) was consistently used in all generation and editing tasks.

\end{document}